\documentclass[twocolumn, switch]{article} 

\usepackage{preprint}

\usepackage{enumitem}
\usepackage{amsmath, amsthm, amssymb, amsfonts}
\usepackage{bm}
\usepackage{amsmath}
\usepackage{graphicx}
\usepackage{subfigure}
\usepackage[algoruled,algo2e]{algorithm2e}

\usepackage{setspace}
\usepackage{amssymb}
\usepackage{booktabs}
\usepackage{array}
\usepackage{color}
\usepackage{multirow}
\usepackage{multicol}
\usepackage{threeparttable}
\usepackage{url}
\usepackage[page]{appendix}
\usepackage{times}
\usepackage{epsfig}
\usepackage{graphicx}
\usepackage{amsmath}

\usepackage{bbding}
\usepackage{booktabs}
\usepackage{mathtools}
\usepackage{color}
\usepackage{threeparttable}
\usepackage{float}

\usepackage[numbers,square]{natbib}
\usepackage[utf8]{inputenc}	
\usepackage[T1]{fontenc}	
\usepackage{xcolor}		
\usepackage[colorlinks = true,
            linkcolor = blue,
            urlcolor  = blue,
            citecolor = blue,
            anchorcolor = black]{hyperref}	
\usepackage{booktabs} 		
\usepackage{nicefrac}		
\usepackage{microtype}		
\usepackage{lineno}		
\usepackage{float}			

\usepackage{newfloat}
\DeclareFloatingEnvironment[name={Supplementary Figure}]{suppfigure}
\usepackage{sidecap}
\sidecaptionvpos{figure}{c}

\usepackage{titlesec}
\titlespacing\section{0pt}{12pt plus 3pt minus 3pt}{1pt plus 1pt minus 1pt}
\titlespacing\subsection{0pt}{10pt plus 3pt minus 3pt}{1pt plus 1pt minus 1pt}
\titlespacing\subsubsection{0pt}{8pt plus 3pt minus 3pt}{1pt plus 1pt minus 1pt}

\usepackage{graphics}
\usepackage{hyperref}
\usepackage{color}
\usepackage{multirow}
\usepackage{hhline}
\usepackage{subfigure}
\usepackage{lipsum}

\title{Learning Knowledge Representation with Meta Knowledge Distillation for Single Image Super-Resolution}

\usepackage{authblk}

\author{Han Zhu, Zhenzhong Chen, Shan Liu}

\begin{document}
\pagestyle{empty}

\twocolumn[ 
  \begin{@twocolumnfalse} 
  
\maketitle

\begin{abstract}

Although the deep CNN-based super-resolution methods have achieved outstanding performance, their memory cost and computational complexity severely limit their practical employment. Knowledge distillation (KD), which can efficiently transfer knowledge from a cumbersome network (teacher) to a compact network (student), has demonstrated its advantages in some computer vision applications. The representation of knowledge is vital for knowledge transferring and student learning, which is generally defined in hand-crafted manners or uses the intermediate features directly. In this paper, we propose a model-agnostic meta knowledge distillation method under the teacher-student architecture for the single image super-resolution task. It provides a more flexible and accurate way to help the teachers transmit knowledge in accordance with the abilities of students via knowledge representation networks (KRNets) with learnable parameters. In order to improve the perception ability of knowledge representation to students' requirements, we propose to solve the transformation process from intermediate outputs to transferred knowledge by employing the student features and the correlation between teacher and student in the KRNets. Specifically, the texture-aware dynamic kernels are generated and then extract texture features to be improved and the corresponding teacher guidance so as to decompose the distillation problem into texture-wise supervision for further promoting the recovery quality of high-frequency details.
In addition, the KRNets are optimized in a meta-learning manner to ensure the knowledge transferring and the student learning are beneficial to improving the reconstructed quality of the student. Experiments conducted on various single image super-resolution datasets demonstrate that our proposed method outperforms existing defined knowledge representation related distillation methods, and can help super-resolution algorithms achieve better reconstruction quality without introducing any inference complexity.

\end{abstract}

\vspace{0.4cm}

  \end{@twocolumnfalse} 
] 

\newcommand\blfootnote[1]{%
\begingroup
\renewcommand\thefootnote{}\footnote{#1}%
\addtocounter{footnote}{-1}%
\endgroup
}

{\blfootnote{Corresponding author: Zhenzhong Chen, E-mail:zzchen@ieee.org}}

\section{INTRODUCTION}

Single image super-resolution (SISR) is a fundamental vision task of reconstructing a high-resolution (HR) image from its low-resolution (LR) counterpart. In addition to improving image resolution and quality, it could be beneficial in many computer vision applications, such as face recognition~\cite{FR_1, FR_2} and person re-identification~\cite{RID_1}. To sum up, SISR has obtained a lot of attention.

With the development of deep learning technology, convolutional neural networks (CNNs) has been widely used in SISR to learn how to generate a high-resolution image from a low-resolution image and has achieved significant gain over the traditional methods, and have become the mainstream methods to deal with the ill-posed problem. Many experimental results indicate that deepening the depth and enlarging the width of networks can efficiently improve the reconstruction quality. 
Consequently, high computation complexity and large memory cost severely impact the wide application of these cumbersome networks in practice. On the contrary, lightweight networks have apparent advantages in practical deployment, but their performance is limited. Hence, we expect to achieve a better tradeoff between accuracy and complexity by improving the behavior of lightweight networks no matter how the architectures are designed. 

Knowledge distillation (KD) is one of the standard model compression methods. Since it can be flexibly applied to various network frameworks and is less dependent on hardware, KD has been thoroughly studied and has shown good performance in promoting the accuracy of lightweight networks (student) via transferring knowledge from the high-accuracy but cumbersome networks (teacher). Related researches mainly focus on some typical high-level vision computer vision tasks, such as classification~\cite{hinton,resKD}, semantic segmentation~\cite{SKD}, action recognition~\cite{SKD_AR,semKD_AR} and so on. Considering the differences in applications, knowledge distillation for low-level vision tasks also deserves to be investigated.

Therefore, we study the intermediate feature knowledge distillation strategy under teacher-student architecture for classical PSNR-oriented single super-resolution task in this paper. Although the SR study is based on bicubic degradation, it still attracts the attention of many researchers and has a reference and inspiration for research such as realSR. The representation of knowledge is a research branch of KD and is vital for knowledge transfer and student learning. Some of the existing knowledge representations are usually in hand-crafted manners. For example, AT~\cite{AT} defines the knowledge as attention maps from intermediate features, and FAKD~\cite{SA} aims at the SISR task to explore the spatial-affinity matrix to make students mimic their teachers. In brief, these methods artificially define the knowledge representation and determine whether to discard or retain information by expert knowledge. Even if they achieve a pronounced distillation effect, they may lead to suboptimal solutions, potentially discarding helpful information when the knowledge required for the student is inaccurately described or formulated. Besides, different students may need different guidance due to differences in abilities and receptive fields and should not be generalized in terms of knowledge expression. 

Different from the above human-driven or experience-driven forms of knowledge representation, some studies directly use the intermediate features of teachers as their knowledge representations, and then use learnable parameters to make students imitate teachers, thereby avoiding important information being discarded. In addition, in order to improving the distillation efficiency, some works~\cite{VID,hegde2020variational,tian2019contrastive} try to explain why knowledge distillation works from the perspective of information theory and formulate the distillation process as maximizing the mutual information between the representations of teacher and student.
Typically, VID~\cite{VID} defines a variational distribution (e.g., Gaussian or Laplace distribution) with learnable mean and variance to approximate the true conditional distribution. However, there are some problems with these methods when employed in the SISR task. Firstly, the true distribution is unknown and intractable, which means there will always be parts of information that cannot be comprehended and followed by the student. Then, in the SISR task, there is a high correlation between student and teacher reconstruction results and even the input low-resolution images. If the knowledge is transferred without screening, the maximization of mutual information potentially mainly focuses on the redundant low-frequency information while ignoring the guidance that students need to restore the high-frequency related details.

In summary, learning the knowledge representation of super-resolution network distillation mainly focuses on dealing with correlations and dependencies between teachers and students with different abilities and requirements, and then accurately describing and transferring the needed information for SISR to improve the distillation efficiency. 

In this paper, aiming at the above problems, we propose a model-agnostic meta knowledge distillation method for the single image super-resolution task. First, we design KRNets with learnable parameters to convert intermediate features into transferred knowledge. Unlike previous works, KRNets take improving the students' reconstruction quality as the optimization target through meta-learning, rather than just imitating teachers. Secondly, considering that the key to the super-resolution task lies in high-frequency texture recovery, in KRNets, the texture-aware kernels are generated from the intermediate features of students. On the one hand, the kernels can extract useful texture-related information as transferred knowledge. On the other hand, considering that students with different abilities need different guidance, the convolution kernels obtained from the students' feature maps can adjust knowledge representations according to their requirements and help teachers teach students according to their aptitude.
The proposed method can be applied to various DNN architectures. The main contributions can be summarized as follows:

\begin{itemize}
	\item A model-agnostic knowledge distillation method is proposed for single image super-resolution task. It provides a more flexible and accurate way to determine the representation of transferred knowledge via the proposed KRNets with learnable parameters, and achieves superior performance in promoting the reconstruction quality of students.
	\item KRNets introduces the dynamic kernels completely generated from the student features modified by the correlation between the teacher and student, to transform the intermediate outputs into the transferred knowledge in a meta-learning manner. It can improve the perception ability of knowledge representation to SR results and boost the distillation efficiency.  
	\item Texture-aware dynamic kernels are proposed to extract texture features to be improved and the corresponding teacher guidance from complex intermediate outputs. These descriptors can decompose the distillation problem into texture-wise supervision to further promote the recovery quality of high-frequency details. 

\end{itemize}

The rest of the paper is organized as follows. Section~\ref{subsec:related_work} reviews the deep learning based SR methods and knowledge distillation. Section~\ref{subsec:proposed_method} elaborates the proposed meta knowledge distillation method in detail. Section~\ref{subsec:experiments} evaluates and analyzes experimental results. Finally, Section~\ref{sec:conclusion} summarizes our work.

\begin{figure*}[t]
	\begin{center}
		\includegraphics[width=0.95\linewidth]{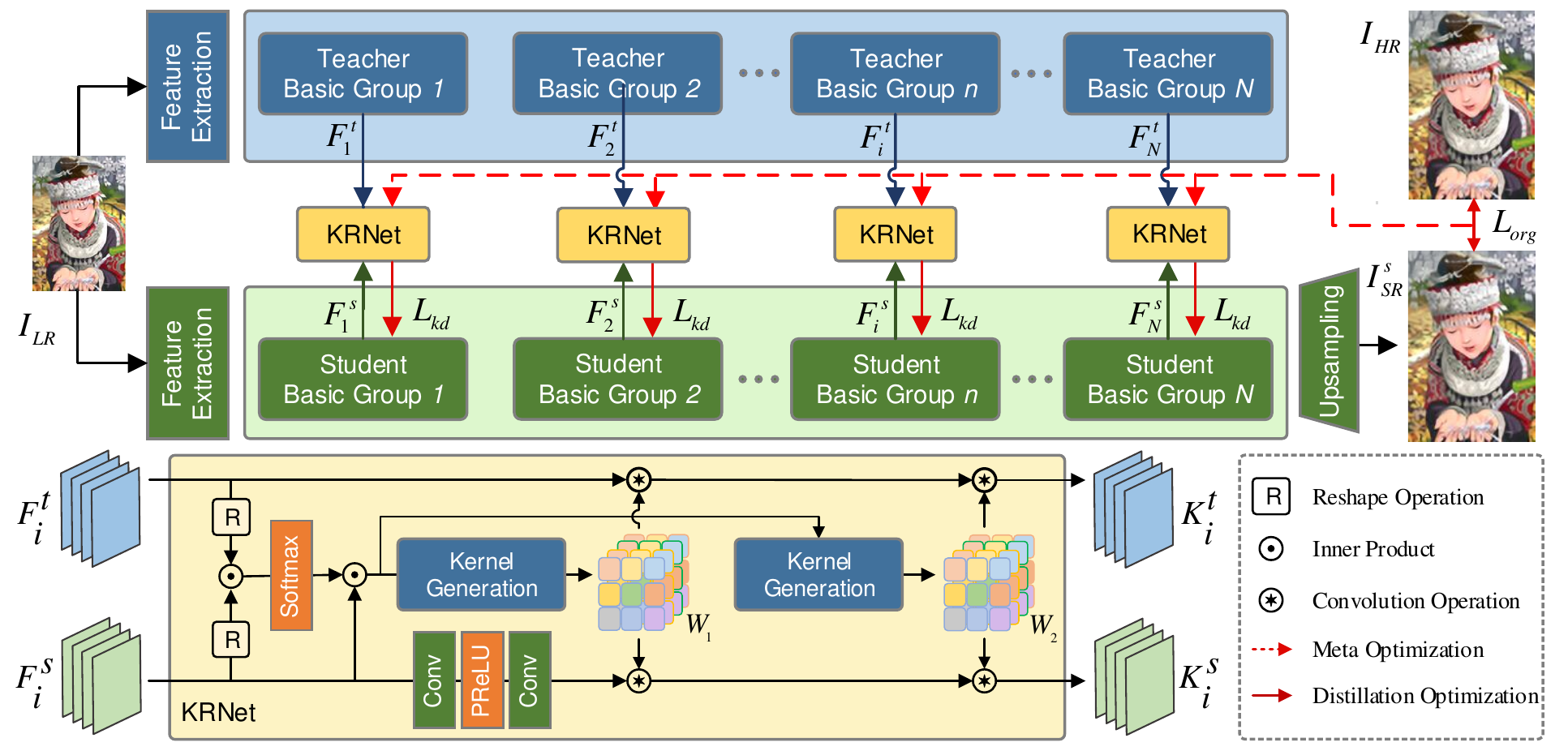}
	\end{center}
	\caption{The framework of proposed meta knowledge distillation for the SISR task. During training, the parameters in the teacher network are fixed, when the student network, meta networks and the discriminator will be optimized.
	}
	\label{fig:framework}
\end{figure*}

\section{Related Work}
\label{subsec:related_work}
Here we review deep learning based SISR methods and knowledge distillation respectively.

\textbf{Single image super-resolution}. SISR has many research branches, such as the classical bicubic degradation SR (including PSNR-oriented and perception-driven~\cite{SRGAN,ESRGAN}), real SR~\cite{realSR,realSR_qa}, reference-based SR~\cite{refSR_zheng,refSR_sun} and so on. This paper mainly focuses on PSNR-oriented algorithms.

SRCNN~\cite{SRCNN} is the first one to introduce deep learning into the SISR task. The majority aim at high PSNR and usually employ L1 or MSE loss as the loss function. For example, SRDenseNet~\cite{SRDenseNet} deepens the depth of network and employs dense blocks to improve the reconstruction quality. 
EDSR~\cite{EDSR} removes the batch normalization layers based on SRResNet~\cite{EDSR} and expands the width and depth of models with the stable training. RDN~\cite{RDN} combines dense connection and residual learning, as well as applies local feature fusion and local residual learning.
With the development of attention mechanisms, many studies have introduced attention blocks to improve the accuracy of super-score reconstruction. For instance, RCAN~\cite{RCAN} employs the channel-wise attention and designs a residual in residual architecture. SAN~\cite{SAN} and HAN~\cite{HAN} adopt the second-order channel attention and holistic layer attention separately. For better use of global context information, CSNLN~\cite{CSNLN} proposes the cross-scale non-local attention and combines local, in-scale non-local, and cross-scale non-local feature correlations to embrace rich external statistics. NLSN~\cite{NLSN} enforces sparsity in non-local operation for SISR task via a novel non-local sparse attention module. Inspired by context-gated convolution~\cite{context_gated}, CRAN~\cite{CRAN} can adaptively modulate the convolution kernel according to the global context enhanced by semantic reasoning. To generate more accurate textures and improve subjective quality, SeaNet~\cite{SeaNet} employs a soft-edge network to help image reconstruction, while TDPN~\cite{TDPN} focuses on local region feature recovery and textures preservation.

\textbf{Knowledge distillation}. Knowledge distillation is one of the model compression approaches. It promotes the accuracy of lightweight networks (student) via transferring knowledge from high-accuracy but cumbersome networks (teacher). Recent research~\cite{VID, hegde2020variational, tian2019contrastive} formulates KD as maximizing the mutual information between the representations of teacher and student. 
The representation of knowledge is vital to the transfer of knowledge and learning of the student network. 
The majority mainly focus on high-level vision tasks, such as image classification~\cite{Fitnet, AT, L2T-ww}, structured prediction~\cite{SKD} and so on. According to the forms of distilled knowledge, it can be divided into response-based, feature-based and relation-based distillation approaches. 

Response or logistic-based knowledge~\cite{hinton} usually refers to the neural response of soft logistic label (the last output layers) of the teacher models. 
Therefore, it is not suitable for being applied to low-level vision tasks. Both feature-based KD and relation-based KD methods rely on intermediate-level supervision from the teacher model. For example, Fitnet~\cite{Fitnet} uses feature representations as hints to improve the reconstruction quality of students. AT~\cite{AT} defines the knowledge as spatial attention map. FSP~\cite{FSP} explores the relationships between different layers via the flow of solution procedure matrix. Although feature-based and relation-based KD can deal with cross domain transfer and low-level vision tasks, knowledge distillation methods for super-resolution tasks, especially the representations of knowledge, still have a lot of space for exploration. FAKD~\cite{SA} proposes feature affinity-based Knowledge, while Huang \textit{et al.}~\cite{PC} define the transferred knowledge as the pixel-correlation for SR network binarization. PISR~\cite{PISR} introduces privileged information to train specific teachers and conducts distillation by adopting VID~\cite{VID}. PAMS\cite{PAMS} introduces a structured knowledge transfer (SKT) loss to fine-tune the proposed quantized network. Unlike these approaches based on the classic teacher-student framework, Zhang \textit{et al.}~\cite{data-free} propose a data-free knowledge distillation framework for super-resolution when lacking of original training dataset, and CSD~\cite{CSD} employs contrastive learning to perform self-distillation.

Since this paper adopts meta-learning, we next review the application of meta-learning in knowledge distillation studies. Contrary to conventional machine learning that uses fixed learning algorithms (inner or base algorithms) trained from datasets to solve corresponding tasks, meta-learning, also known as learning to learn, aims to improve the algorithms (outer or meta algorithms) itself~\cite{meta_survey}.  
During meta-learning, an outer (or upper/meta) algorithm improves the outer objective by updating the inner learning algorithm. The objective could be the performance or the learning speed of the inner algorithm generally.
According to the definition of meta-learning, many details requiring to be exhaustively tuned in conventional algorithms, such as hyper-parameters, can be re-defined as a meta-learning problem. 
L2T-ww~\cite{L2T-ww} designs meta-transfer networks that can determine what and where to transfer. Flennerhag \textit{et al.}~\cite{Flennerhag} propose a lightweight framework for meta-learning over task manifolds. Peng \textit{et al.}~\cite{Peng} designs a new transfer architecture for few-shot image recognition. Liu \textit{et al.}~\cite{Liu} propose the semantic-aware knowledge preservation method for image retrieval. Zhou \textit{et al.}~\cite{bert} adjust teacher networks' parameters to provide better meta-learning guidance.
It is worth noting that although it seems that both L2T-ww~\cite{L2T-ww} and the proposed method in this paper have studied the problem of what to transfer, L2T-ww adopts the same knowledge representation definition and optimization method as FitNet~\cite{Fitnet}. Meta-learning is employed to automatically adjust the weights of different features distillation loss among all student and teacher CNN kernels. In brief, the knowledge representation in L2T-ww is weighted intermediate features.
To our best knowledge, few works have been proposed to employ meta-learning to explore how to automatically determine the representation of knowledge for low-level vision tasks, especially for texture recovery in SISR task.

\section{Proposed Method}
\label{subsec:proposed_method}

Although the architectures of existing SISR models have their own characteristics, many typical state-of-the-art algorithms employ a post-upsampling framework, where upsampling operation converts $f_{LR}$, the feature maps extracted from input image $I_{LR}$, into the reconstructed image $I_{SR}$. Generally, the feature extraction stage consists of several basic convolutional blocks. Therefore, the paper takes the common architecture as an example to expound on the proposed model-agnostic knowledge distillation. The proposed meta knowledge distillation method and the optimization processing will be explained detailedly in this section.

\subsection{Problem definition}
Since SISR is a typical low-level vision task, we apply an intermediate-level supervision knowledge distillation strategy to transfer knowledge from a high-accuracy but cumbersome network to a lightweight network $f_{\theta}$ with parameters $\theta$. The general T-S architecture is illustrated in Figure~\ref{fig:framework}. The teacher and student networks can be divided into $N$ basic groups, respectively. For the convenience of the display, $N$ is equal to 4 in the figure. The feature of the $i$-th teacher basic group is denoted as $F_i^t$ with $c^t$ channels, and the feature of basic student group is $F_i^s$ with $c^s$ channels. Generally, a basic group of the teacher network contains more basic blocks than that of the student network, and the value of $c_t$ is larger than $c_s$. 

The study of intermediate-level distillation can be tracked back to FitNet~\cite{Fitnet}, which defines the output of teacher's hidden layers as hints and instructs the student to mimic the teacher by minimizing the gap before hints and guided (output of student's hidden layers). Inspired by the work, there have been many endeavors to study how to match the hints and the guided layers via transformation of hints and guided feaures~\cite{KD_survey}. The distillation loss can be defined as:
\begin{equation}
\begin{aligned}
\mathcal{L}_{kd} = D(R_t(F^t), R_s(F^s))
\end{aligned}
\end{equation}
where $R_t(\cdot)$ and $R_s(\cdot)$ are functions transforming hints and guided features to knowledge representation respectively. $D(\cdot)$ is the distance measuring the similarity between hints and guided features. 

Knowledge representation directly affects how teachers teach and how students learn. As mentioned above, the definition of knowledge representation is roughly divided into two directions. One is to directly use the hidden layer outputs of the teacher as the knowledge to be taught, so that students can completely use the teacher as a benchmark to imitate by maximizing their mutual information, that is, teacher-driven. The other is to artificially define the transformation form of the hidden features according to experience, that is, human-driven. However, these methods do not consider the aptitude of students. Limited by receptive field and learning ability, students cannot fully digest and imitate all the knowledge and guidance from teachers, thus affecting the effect of knowledge distillation. Especially in the SISR task, the reconstruction quality depends on the acquisition of inter-pixel relationships and high-frequency information, and there is a high similarity between LR and HR. Hence, it is necessary to explore a more flexible way of determining the knowledge representation, which is supposed to take the needs of students into consideration and teach students in accordance with their aptitude.

\subsection{Meta-learning based knowledge representation}
\label{subsec:labelone}

Different from hand-crafted manners, we design knowledge representation networks (KRNets) $\mathcal{M}_{\phi}$ with learnable parameters $\phi$, and meet the above requirements from parameters update and the network architecture design. 

\textbf{Meta optimization.} Contrary to conventional machine learning, which uses fixed learning algorithms to solve corresponding tasks, meta-learning aims to improve the algorithms themselves. An outer algorithm improves the outer objective by updating the inner learning algorithm during meta-learning. Our goal is to encourage KRNets to be optimized in a direction that enables the student to achieve higher reconstruction quality when minimizing the distillation loss $\mathcal{L}_{kd}$. Taking the training processing of student $f_{\theta}$ as the inner loop, the general bilevel scheme is used as follows:
\begin{itemize}
	\item Minimizing the inner objective $\mathcal{L}_{kd}$ to update $\theta$ for $T$ times.
	\item Update $\phi$ to minimize the meta objective $\mathcal{L}_{org}$.
\end{itemize}

In the first step, endowing the current parameter $\theta_0=\theta$, we optimize the student for $T$ times via minimizing the distillation loss $\mathcal{L}_{kd}$. Then, $\mathcal{L}_{org}$ of $\theta_T$ is calculated to compute the impact of KRNets and $\mathcal{L}_{kd}$ on student performance during the first step, thereby updating $\phi$ by minimizing $\mathcal{L}_{org}$ as the red dotted line in Figure~\ref{fig:framework} shows. It measures the gap between SR results of students and the corresponding HRs and adopts L1 loss in this paper. Although there is no direct gradient correlation between $\phi$ and $\mathcal{L}_{org}$ at $t=T$, $\theta_T$ is obtained from $\theta_0$ by minimizing $\mathcal{L}_{kd}$, which involves $\phi$. For convenience, SGD is used here to sort out the gradient calculation process of $\phi$ at $t=T$, while ADAM is adopted for model training in the experiments.
\begin{equation}
\theta_t = \theta_{t-1} - \alpha\nabla_{\theta_{t-1}}\mathcal{L}_{kd}^{t-1}
\end{equation}
\begin{equation}
\begin{aligned}
\nabla_{\phi}\mathcal{L}_{org} &= \nabla_{f}\mathcal{L}_{org}\nabla_{\theta_T}f_{\theta_T}\nabla_{\phi}\theta_T \\
&= \nabla_{f}\mathcal{L}_{org}\nabla_{\theta_T}f_{\theta_T}\nabla_{\phi}[\theta_{T-1}-\alpha\nabla_{\theta_{T-1}}\mathbb{E}[\mathcal{L}_{kd}^{T-1}]]
\end{aligned}
\end{equation}

It is worth noting that since the student $f_{\theta}$ is optimized by minimizing the distillation loss $\mathcal{L}_{kd}$ in the inner loop, the convergence of $\mathcal{L}_{kd}$ needs to be maintained when updating the parameters $\phi$. In summary, the training of KRNets $\mathcal{M}_{\phi}$ relies on calculating the gradient of $\theta$'s gradient to update the parameters so as to achieve the purpose of helping to improve the accuracy of students via knowledge distillation. Hence, all parameters in $\mathcal{M}_{\phi}$ need to be directly related to the student $f_{\theta}$.

\begin{figure}[t]
	\begin{center}
		\includegraphics[width=1\linewidth]{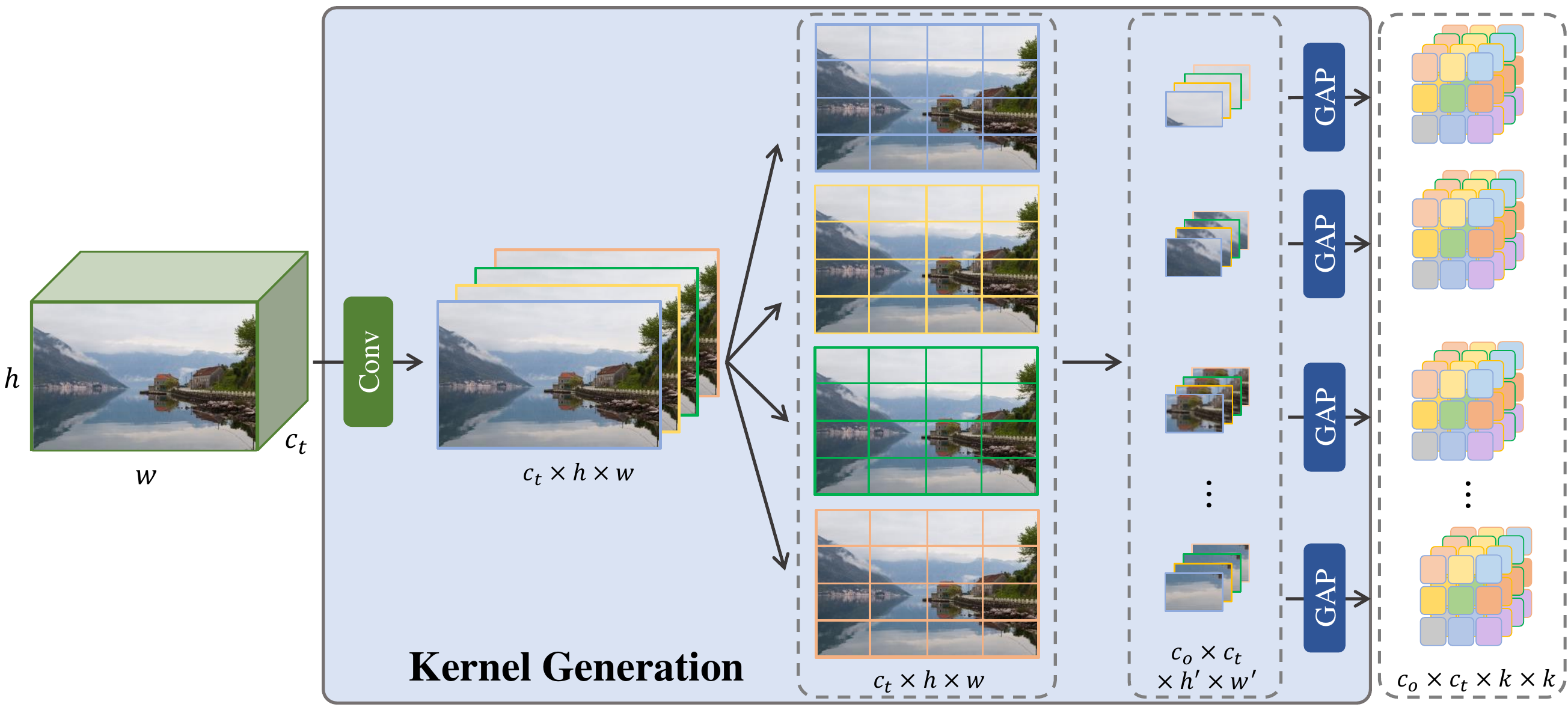}
	\end{center}
	\caption{A brief view of the proposed texture-aware convolution kernel generation. }
	\label{fig:kernel}
\end{figure}

\textbf{Knowledge representation networks.} As discussed earlier, transferred knowledge should be in accordance with students' aptitude so that they can acquire knowledge and improve reconstruction accuracy. Otherwise, knowledge beyond their cognitive ability will be a fantasy for students. We propose the KRNets to formulate the functions transforming hints and guided features to knowledge representation through convolution kernels. In addition to that, meta-learning can adjust the optimization direction of KRNet according to the current gap between $\mathcal{I}_{SR}^s$ and $\mathcal{I}_{HR}$. We use the hidden layer outputs of students to generate the dynamic convolution kernel for further improving the sensitivity of knowledge representation to students' abilities. At the same time, this solution also meets the needs of meta optimization for gradient transmission and calculation. The architecture of the proposed KRNet is shown in Figure~\ref{fig:framework}.

First, the student's feature maps $F^s_i\in\mathbb{R}^{c_s\times h\times w}$ are processed to $F'^s_i\in\mathbb{R}^{c_t\times h\times w}$ by two $1\times1$ convolutions with an activation layer in the middle, so that the number of channels matches that of the teacher $F^t_i\in\mathbb{R}^{c_t\times h\times w}$. Then employ the generated convolution kernels to perform convolution operations on $F^t_i$ and $F'^s_i$ to obtain the knowledge representation of teacher and student, respectively. The distance between $K^s_i$ and $K^t_i$ is adopted as the distillation loss $\mathcal{L}_{kd}^i$. In addition to considering students' aptitude, we also adjust $F^s_i$ according to the teacher's ability to boost distillation efficiency. The shape of $F^s_i$ and $F^t_i$ are converted into $c_s\times hw$ and $c_t\times hw$, respectively. Then, inner product and softmax operations are adopted to acquire the correlation on the channel between the feature maps extracted by the student and the teacher. Finally, the correlation can dynamically modify $F^s_i$ via the inner product.

The processing procedure of KRNets can be expressed as follows:
\begin{equation}
W = G(\sigma(F_i^s\odot F_i^t)\odot F_i^s) \label{con:E4}
\end{equation}
where $G(\cdot)$ denotes the kernel generation block, which will be expounded later. $\odot$ and $\sigma(\cdot)$ mean the inner product and softmax operations. Both $W_1\in\mathbb{R}^{c_o\times c_t \times k\times k}$ and $W_2\in\mathbb{R}^{c_o\times c_t \times k\times k}$ are obtained by Equation~\ref{con:E4}, but have different weights in the kernel generation blocks.

\begin{equation}
K^s_i = act(DM(F_i^s)\otimes W_1)\otimes W_2
\end{equation}
\begin{equation}
K^t_i = act(F_i^t\otimes W_1)\otimes W_2
\end{equation}
where $\otimes$ and $act(\cdot)$ denote the convolution and activation operations. $DM(\cdot)$ is the dimension matching process described above. The total distillation loss is formulated as:
\begin{equation}
\mathcal{L}_{kd} = \sum_{i=0}^N\vert K^s_i-K^t_i\vert
\end{equation}

\textbf{Texture-aware dynamic kernel generation.} In the SISR task, the information lost in the process from HR degradation to LR is recovered by acquiring the inter-pixel relationships. It is no doubt that long-term and global information has obvious benefits to improving the reconstruction quality of cumbersome networks, which has been studied and proved in previous work~\cite{CSNLN, CRAN}. However, for a lightweight network, the size of the model and the receptive field directly limit the ability of the network to obtain long-distance information. The pixels that show the most significant contribution to the reconstruction quality of a local area in Gu \textit{et al.}~\cite{LAM} are still the pixels within the short-term range. Therefore, the convolution kernel generation for knowledge extraction still focuses on local texture information.

Procedure of the proposed texture-aware kernel generation is displayed in Figure~\ref{fig:kernel}. Firstly, a $1\times 1$ convolutional layer is employed to reorganize the channel information of $F''_i\in\mathbb{R}^{c_t\times h\times w}$ to prepare for the generation of the kernels. Different from existing dynamic kernel methods~\cite{context_gated, dynamic_conv} modifying the static convolutions among global spatial information, we think that different textures in the super-resolution problem should be treated with different weights and thus propose the texture-aware kernel. In order to simplify KRNets and improve the stability as well as the training speed of meta-learning, we directly segment the feature maps in the spatial dimension to obtain $c_o=K*K$ subpatches with size $c_t\times h'\times w'$, that is, it is assumed that the texture does not change too drastically on each subpatch. Finally, the texture-aware convolution kernel $W\in\mathbb{R}^{c_o\times c_t\times k\times k}$ is generated according to the global information of each subpatches through global adaptive pooling. These texture descriptors extract texture-related features from the intermediate outputs of the students and the teacher as the transferred knowledge, and compute the their gap as the distillation loss $\mathcal{L}_{kd}$.

In summary, we propose a more flexible and accurate way to determine the knowledge representation. The flexibility and accuracy lie not only in updating parameters of the KRNets through meta-learning but also in decomposing the distillation problem into texture-wise supervision and adjusting the knowledge representation according to requirements of students during the training process to achieve the purpose of teaching students according to their aptitude.

\section{Experiments}
\label{subsec:experiments}

\subsection{Datasets and implementation details}
\label{subsec:labeltwo}
We train our networks on DIV2K \cite{DIV2K} dataset. DIV2K is a widely recognized high-resolution image dataset, which consists of 800 images for training and 100 validation images. All low-resolution images applied in our experiments are generated by BI degeneration from corresponding high-resolution images. Performance comparison experiments are conducted on super-resolution benchmarks (Set5~\cite{Set5}, Set14~\cite{Set14}, BSD100~\cite{B100}, and Urban~\cite{Urban100}). The peak noise-signal ratio (PSNR) and structural similarity index measure (SSIM) are employed to evaluate the super-resolution performance on the Y channel of images represented in the YCbCr color space.

Since the proposed meta knowledge distillation (MKD) method is model-agnostic, in addition to conducting ablation and comparison experiments on RCAN~\cite{RCAN}, we also employed some classic SOTA algorithms to verify the effectiveness of the proposed MKD. Furthermore, we set different complexity for student networks to analyze the performance of knowledge distillation algorithms on students with different abilities. Lightweight models are obtained by reducing the depth and width of large models provided by their papers. The degree of information loss is discrepant from different down-sampling scaling factors. To better demonstrate that the proposed method can alleviate undesired distortion, e.g., blur and texture inconsistency, which is more likely to occur in the results of lightweight models, we conduct experiments on $\times 4$ scaling factor. It needs to be emphasized that the proposed method can be applied to any scale factor supported by the original SISR methods. The position of distillation, that is, the number of basic groups $N$ of the student and teacher networks, is settled as 4 in the following experiments.

During training, the mini-batch size is set as 16, and RGB input images are cropped to $192\times 192$ high-resolution patches with corresponding $48 \times 48$ low-resolution patches randomly. The compact networks $T$ are obtained from official public code or trained by following the corresponding descriptions. All of student networks are optimized by Adam~\cite{adam} with $\beta_1=0.9$ and $\beta_2=0.999$. The initial learning rate is set as 1$e$-4 and reduced to 5$e$-6 by CosineAnnealing~\cite{sgdr}. The epochs are adjusted according to the students' model size to ensure convergence and the fairness of comparison.

\subsection{Ablation study}

The ablation study explores and analyzes the proposed knowledge representation and texture-aware kernel generation. The experiments are conducted on the architecture of RCAN~\cite{RCAN}. The teacher and three student networks configuration is shown in Table~\ref{tab:config}. The Flops(floating-point operations per second) is calculated on an LR image of size $512\times512$. 

\begin{table}[]
	\centering
	\caption{The teacher and student networks configuration of RCAN~\cite{RCAN}.}
	\label{tab:config}
	\setlength{\tabcolsep}{2mm}{
		\begin{small}
			\begin{tabular}{c|cc|ccc}
				\toprule[1pt]
				Network & Param.	   & Flops & n\_RGs & n\_RBs & n\_feats \\
				\midrule[1pt]
				RCAN-T  & 15.6M        & 4183.36G &  10    &  20    &  64      \\
				\midrule[0.8pt]
				RCAN-A  & 49.8K        & 22.03G   &  4     &   1    &  16      \\
				RCAN-B  & 106.1K        & 36.68G &  4     &   4    &  16      \\
				RCAN-C  & 420.5K        & 142.19G &  4     &   4    &  32      \\
				\bottomrule[1pt]
			\end{tabular}
		\end{small}
	}
\end{table}

\begin{table*}[!h]
	\centering
	\caption{The ablation results (PSNR / SSIM) of the proposed method. Note: Red color indicates the best performance and Blue color represents the second.}
	\label{tab:ablation}
	\setlength{\tabcolsep}{2mm}{
		\begin{small}
			\begin{tabular}{c|c|ccc|cccc}
				\toprule[1pt]
				Network	& Configuration & is\_meta & KR of T & KR of S & Set5~\cite{Set5} & Set14~\cite{Set14} & BSD100~\cite{B100} & Urban100~\cite{Urban100} \\ 
				\midrule[0.8pt]
				\multirow{5}{*}{RCAN-A}	& Vanilla &  \XSolidBrush      &  -      &  -      & 30.71 / 0.8658  & 27.67 / 0.7562 & 26.96 / 0.7162 & 24.58 / 0.7300 \\
				& Case 1  &  \Checkmark     & -       & 1$\times$1 Conv.    & 30.80 / 0.8676  & 27.73 / 0.7579 & 27.00 / 0.7172  & 24.65 / 0.7330  \\
				& Case 2  &  \Checkmark     & 3$\times$3 Conv.   & 3$\times$3 Conv.      & 30.80 / 0.8681 & \textcolor{blue}{27.75 / 0.7583}  & \textcolor{blue}{27.01} / 0.7178  & 24.67 / 0.7339 \\
				& Case 3  &  \XSolidBrush   & KRNet   & KRNet  & \textcolor{blue}{30.81 / 0.8681}  & 27.75 / 0.7583  & 27.01 / \textcolor{blue}{0.7178}  & \textcolor{blue}{24.67 / 0.7343} \\
				& Case 4  &  \Checkmark     & KRNet   & KRNet  & \textcolor{red}{30.94 / 0.8701}  & \textcolor{red}{27.82 / 0.7597}  & \textcolor{red}{27.04 / 0.7186}  & \textcolor{red}{24.73 / 0.7364}\\
				\midrule[0.8pt]
				\multirow{5}{*}{RCAN-B}	& Vanilla &  \XSolidBrush      & -       & -       & 31.24 / 0.8752   & 27.99 / 0.7640 & 24.17 / 0.7231 & 24.94 / 0.7459 \\
				& Case 1  &  \Checkmark     & -       & 1$\times$1 Conv.  & 31.34 / 0.8772  & 28.06 / 0.7660  & 27.21 / 0.7244  & 25.04 / 0.7501    \\
				& Case 2  &  \Checkmark     & 3$\times$3 Conv.   & 3$\times$3 Conv. & \textcolor{red}{31.48} / \textcolor{blue}{0.8798}  & \textcolor{blue}{28.14 / 0.7676}  & \textcolor{blue}{27.25} / \textcolor{blue}{0.7258}  & \textcolor{blue}{25.14 / 0.7541} \\
				& Case 3  &  \XSolidBrush   & KRNet   & KRNet   & 31.36 / 0.8778 & 28.08 / 0.7665 & 27.23 / 0.7250 & 25.07 / 0.7518 \\
				& Case 4  &  \Checkmark     & KRNet   & KRNet   & \textcolor{blue}{31.47} / \textcolor{red}{0.8798} & \textcolor{red}{28.16 / 0.7680} & \textcolor{red}{27.27 / 0.7264} & \textcolor{red}{25.19 / 0.7562} \\
				\midrule[0.8pt]
				\multirow{5}{*}{RCAN-C}	& Vanilla &  \XSolidBrush      & -       & -       & 31.77 / 0.8841  & 28.31 / 0.7724  & 27.39 / 0.7305  & 25.51 / 0.7674 \\
				& Case 1  &  \Checkmark     & -       &  1$\times$1 Conv.  & 31.84 / 0.8854  & 28.34 / 0.7733  & 27.41 / 0.7313  & 25.56 / 0.7695  \\
				& Case 2  &  \Checkmark     & 3$\times$3 Conv.   & 3$\times$3 Conv.   & \textcolor{blue}{31.90 / 0.8863}  & 28.38 / 0.7739 & \textcolor{blue}{27.43 / 0.7317}   & \textcolor{blue}{25.62 / 0.7713} \\
				& Case 3  &  \XSolidBrush   & KRNet   & KRNet   & 31.90 / 0.8859  & \textcolor{blue}{28.39 / 0.7741}  & 27.43 / 0.7317  & 25.61 / 0.7706 \\
				& Case 4  &  \Checkmark     & KRNet   & KRNet   & \textcolor{red}{31.96 / 0.8871}  & \textcolor{red}{28.40 / 0.7746} & \textcolor{red}{27.45 / 0.7325} & \textcolor{red}{25.71 / 0.7742} \\
				\bottomrule[1pt]
			\end{tabular}
		\end{small}
	}
\end{table*}

\begin{figure}[t]
	\begin{center}
		\includegraphics[width=1\linewidth]{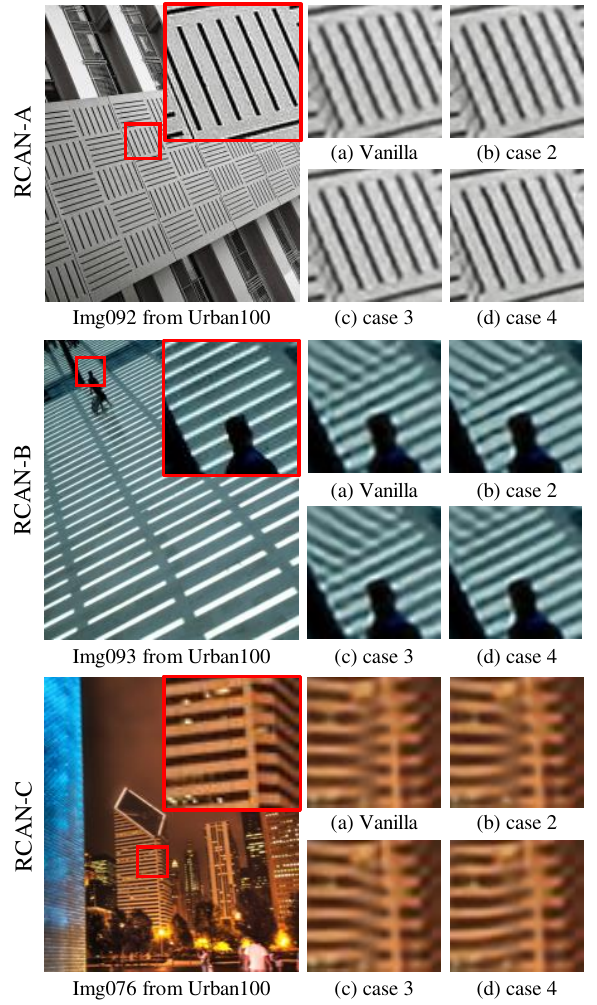}
	\end{center}
	\caption{Visual comparison of different ablation settings. Ground truths are displayed in red boxes.}
	\label{fig:long}
	\label{fig:Ablation_visual}
\end{figure}

\begin{figure}[t]
	\begin{center}
		\includegraphics[width=0.85\linewidth]{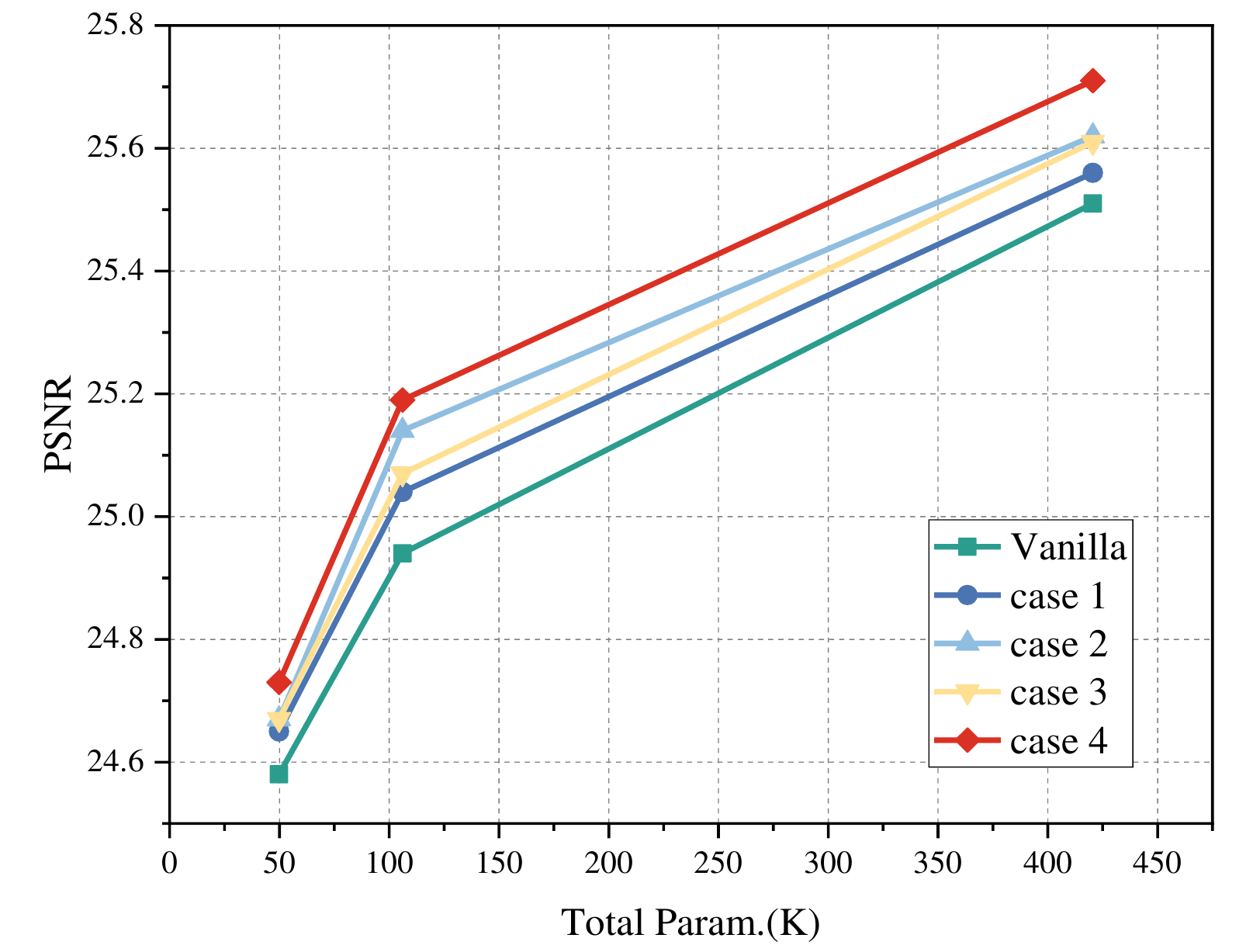}
	\end{center}
	\caption{The relationship between total parameters of different students and the performance for ablation settings on Urban100~\cite{Urban100} dataset.}
	\label{fig:long}
	\label{fig:Ablation_R}
\end{figure}

\subsubsection{Analysis of knowledge representation} We measure the effect of ablation components by enabling them one by one, including whether to use meta-learning to optimize the KRNets, and respective knowledge representations of teacher and student. The results of the different settings are shown in Table~\ref{tab:ablation}, where the check and cross marks indicate that the knowledge representation with learnable parameters is optimized with and without meta-learning, respectively. Vanilla denotes that the model is trained from scratch without knowledge distillation. Case 1 means that the output of the intermediate layers of the teacher is directly regarded as the knowledge to be transferred, while the student uses a $1\times1$ convolution optimized by meta-learning to match the knowledge. Compared with case 1, case 2 uses two $3\times3$ convolutional layers as the knowledge representation of student and teacher, respectively. Case 3 employs the proposed KRNets as the knowledge representation but directly updates the parameters by minimizing the distillation loss $\mathcal{L}_{kd}$. Case 4 is the final proposed solution. The visual comparison of the above ablation settings is displayed in Figure~\ref{fig:Ablation_visual}.

It can be observed that the proposed method (case 4) achieves the best performance. Both the flexibly defined knowledge representation via meta-learning and the texture-aware KRNet bring about significant improvement. Judging from the results of case 1, case 2, and case 4, the form of knowledge representation is essential. Compared with directly requiring students to imitate the teacher's intermediate outputs, rationally screening and integrating features is more conducive to improving the distillation efficiency of SR tasks. It is worth noting that the performance improvement achieved on RCNA-B using two $3\times3$ convolutional layers to transform knowledge (case 2) is significantly higher than that on RCAN-A and RCAN-C. In contrast, the promotions for the three-complexity students of case 4 are more balanced., which can be more clearly observed in Figure~\ref{fig:Ablation_R}. It can be inferred that the proposed KRNet is more sensitive to the reconstruction ability of students so as to cope with the different gaps between teachers and students, that is to say, to achieve teaching according to aptitude. Even without using meta-learning to optimize the learnable parameters (case 3), the KRNet still achieves impressive distillation results. Admittedly, adapting knowledge representations via meta-learning can make knowledge distillation more beneficial to improving student performance.

In Figure~\ref{fig:Ablation_visual}, the reconstruction results of case 4 have sharper and cleaner edges. The proposed method can effectively alleviate problems such as texture dislocation and geometric distortion, which often occur in images with regular and dense textures. For instance, in the third example of the figure, except for case 4, there are obvious wall line dislocations and fractures in other reconstructed results. In addition, beneficial from the KRNet, case 3 has a certain corrective effect on texture restoration compared to vanilla and case 2.

\subsubsection{Analysis of texture-aware kernel generation}

\begin{table}[]
	\centering
	\caption{The PSNR on Urban100~\cite{Urban100} dataset of employing different subpatch numbers $c_0$ in texture-aware kernel generation. Red indicates the best one.}
	\label{tab:kernel}
	\setlength{\tabcolsep}{2mm}{
		\begin{small}
			\begin{tabular}{c|c|ccc}
				\toprule[1pt]
				\multicolumn{2}{c|}{Configuration}     & RCAN-A                       & RCAN-B                       & RCAN-C \\
				\midrule[1pt]
				\multicolumn{2}{c|}{Vanilla}     & 24.58                        & 24.94                        & 25.51  \\
				\multicolumn{2}{c|}{Case 2}       & 24.67                        & 25.14                        & 25.62  \\
				\midrule[0.8pt]
				\multirow{4}{*}{Case 4} &  $c_0 = 4$   & 24.72                        & 25.18                        & 25.66  \\
				& $c_0=16$  & 24.74                        & 25.19                        & 25.68  \\
				& $c_0=64$  & 24.73                        & \textcolor{red}{25.19} & \textcolor{red}{25.71}  \\
				& $c_0=144$ & \textcolor{red}{24.76} & 25.19                        & 25.67  \\
				\bottomrule[1pt]
			\end{tabular}
		\end{small}
	}
\end{table}

\begin{figure*}[t]
	\begin{center}
		\includegraphics[width=1\linewidth]{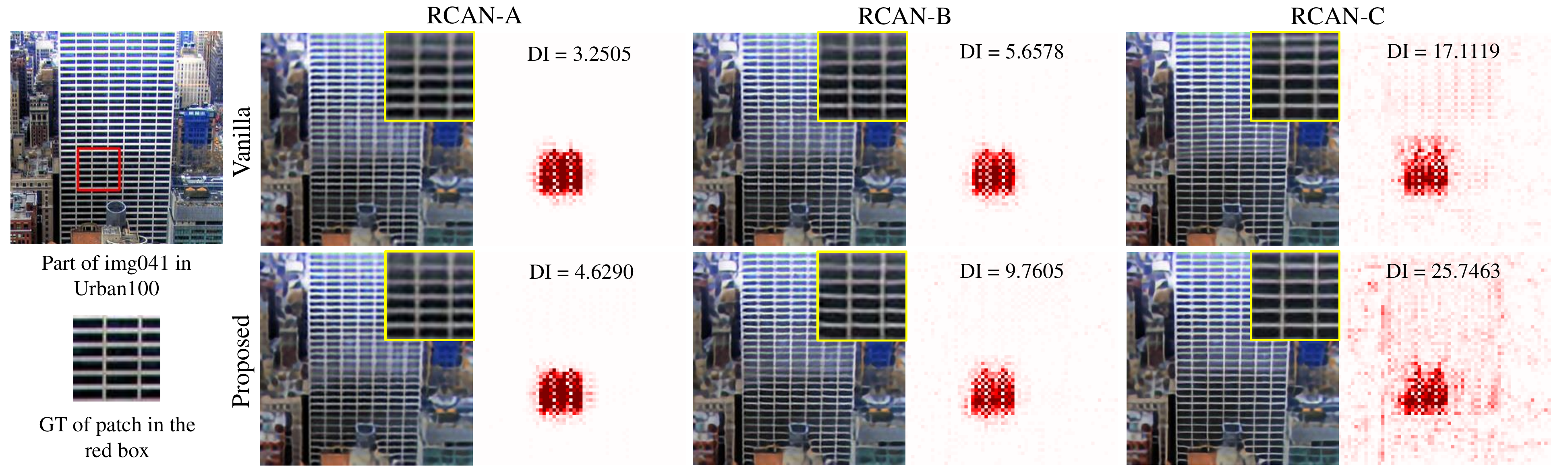}
	\end{center}
	\caption{Visual comparison of the SR images and LAM\cite{LAM} attribution results, which visualize the importance of each pixel to the reconstruction results inside the red box. The corresponding reconstruction results are shown in the yellow boxes. DI indicates the range of involved pixels.}
	\label{fig:long}
	\label{fig:LAM}
\end{figure*}

The above ablation study on knowledge representation has shown that the proposed KRNet is helpful for texture recovery and performance improvement. In this subsection, we explore the effect of the degree of texture decomposition on the distillation effect when generating texture-aware kernels in the KRNet, and further analyze the role of knowledge distillation in improving student performance.

We conduct experiments with four settings for the number of subpatches $c_0$ (defined in section~\ref{subsec:labelone}) in texture-aware kernel generation, and the results are shown in Table~\ref{tab:kernel}. For an input LR patch, the number of subpatch divisions corresponds to the size of the division window, that is, the degree of texture decomposition. For example, $c_0=64$ represents dividing the $48\times48$ input into 64 subpatches with size $8\times8$, and the $3\times3$ convolution kernels are generated from these subpatches, corresponding to the texture inside them. In Table~\ref{tab:kernel}, in the case of employing meta-learning, the proposed kernel has obvious advantages compared to the direct use of normal convolutions in case 2. On the distillation results of student networks with different complexity, the performance of long-distance patch partitioning ($c_0=4$) is worse than others, which indicates that the lightweight SR networks are more focused on capturing local or short-distance pixel relationships for reconstructing details. In contrast, the long-distance context information is more conducive to improving the performance of cumbersome networks. In choosing the optimal number of divisions, students of different complexity have different preferences. For the lightest network RCAN-A, $c_0=144$ performs best, which means that being sensitive to local information benefits its reconstruction quality. But that doesn't mean the more subpatches, the better. For RCAN-B, the value of $c_0$ has no significant effect, and different information ranges can be used to improve performance. For RCAN-C, the most capable network among the three students, $c_0=64$ has apparent advantages. Because RCAN-C has a certain ability to acquire features and integrate information, improving the perception of local textures has a limited effect on improving its capacity. For convenience, all the lightweight networks conducted the proposed KD method in other experiments adopts $c_0=64$.
What needs to be stated again is that there are indeed better ways to generate kernels more suitable for texture decomposition and extraction. However, since meta-learning needs to revisit the optimization results of $T$ times and use the gradient of the gradient to update the parameters of KRNet, adopting a more complex approach to generate kernels will result in an exponential increase in meta-optimization complexity. We adopt patch segmentation to balance performance, complexity, and training stability.

In order to verify and analyze the specific role of KD in improving the performance of lightweight networks, we adopted the tool LAM\footnote{Code(demo): https://x-lowlevel-vision.github.io/lam.html} proposed by Gu et al.~\cite{LAM}, which can visualize the importance of each pixel to the reconstruction of a specified area to compare the SR results with and without the proposed KD method. The visualization results are shown in Figure 7. The darker the red, the more significant contribution of the pixel to the reconstruction of the current frame. DI reflects the range of involved pixels. Although DI cannot be entirely equivalent for accuracy, it can reflect the network's ability to acquire features and integrate information. Due to the existence of global operations in RCAN, the theoretical receptive field of RCAN-A/B/C covers the global area. However, in fact, the pixels that have the most significant impact on the current area are still clustered around. The distillation method proposed in this paper can significantly improve the networks' ability to perceive texture so that they can learn from a wider range of similar texture regions to improve the reconstruction quality. It has been reflected in the significantly increased DI values.

\begin{figure}[t]
	\begin{center}
		\includegraphics[width=0.98\linewidth]{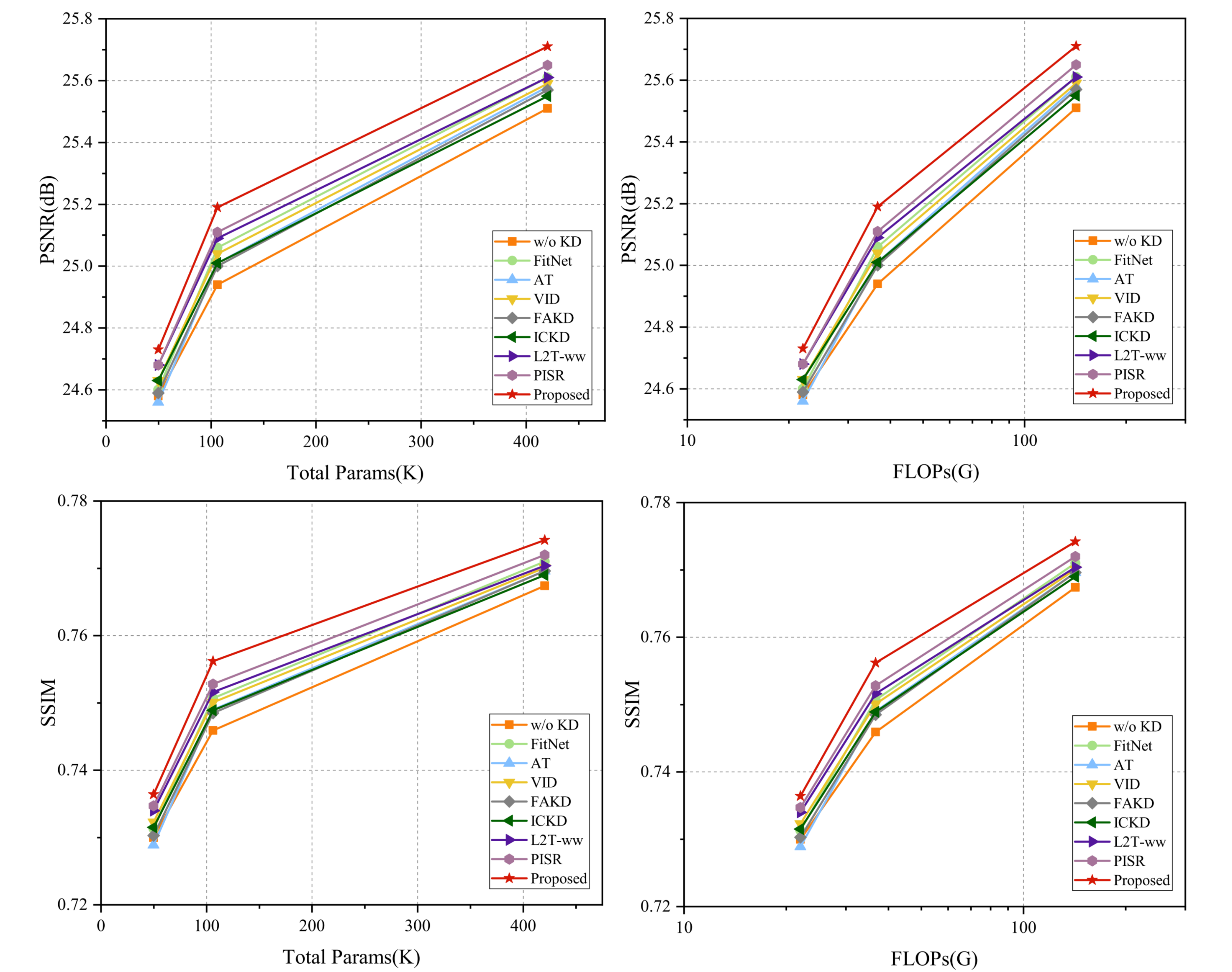}
	\end{center}
	\caption{The relationship between complexity of different students (total parameters and FLOPs) and the performance (PSNR and SSIM) for ablation settings on Urban100~\cite{Urban100} dataset.}
	\label{fig:long}
	\label{fig:Tradeoff}
\end{figure}

\begin{table*}[!h]
	\centering
	\caption{The comparison results with other knowledge distillation methods on SR benchmarks. Red indicates the best performance and blue indicates the second one.}
	\label{tab:analysis}
	\setlength{\tabcolsep}{3mm}{
		\begin{small}
			\begin{threeparttable}
				\begin{tabular}{c|c|cccccccc}
					\toprule[1pt]
					\multirow{2}{*}{Network}	& \multirow{2}{*}{Method}  & \multicolumn{2}{c}{Set5~\cite{Set5}} & \multicolumn{2}{c}{Set14~\cite{Set14}} & \multicolumn{2}{c}{BSD100~\cite{B100}} & \multicolumn{2}{c}{Urban100~\cite{Urban100}}                  \\
					
					& & PSNR   &SSIM    &PSNR   &SSIM   &PSNR   &SSIM   &PSNR   &SSIM             \\
					\midrule[0.8pt]
					\multirow{9}{*}{RCAN-A}& Vanilla     & 30.71 & 0.8658  & 27.67 & 0.7562 & 26.96 & 0.7162  & 24.58  & 0.7300      \\
					& FitNet~\cite{Fitnet}   & 30.70  & 0.8665  & 27.68  & 0.7568  & 26.97  & 0.7168   & 24.60  & 0.7313    \\
					& AT~\cite{AT}   & 30.67  & 0.8648 & 27.65 & 0.7557  & 26.95 & 0.7156  & 24.56  & 0.7289        \\
					& VID~\cite{VID}   & 30.77  & 0.8673 & 27.70  & 0.7571  & 26.98  & 0.7169   & 24.63  & 0.7323                 \\
					& FAKD~\cite{SA}   & 30.72  & 0.8660  & 27.66  & 0.7563  & 26.96  & 0.7165  & 24.59  & 0.7303                \\
					& ICKD~\cite{ICKD} & 30.77 & 0.8665  & 27.69 & 0.7566  & 26.97 & 0.7165  & 24.63 & 0.7315      \\
					& L2T-ww\tnote{*}~\cite{L2T-ww}   & \textcolor{blue}{30.87} & \textcolor{blue}{0.8690} & 27.75 & 0.7582   & \textcolor{blue}{27.02} & \textcolor{blue}{0.7182}  & 24.68 & 0.7340                  \\
					& PISR\tnote{**}~\cite{PISR}   & 30.86 & 0.8688	& \textcolor{blue}{27.78} & \textcolor{blue}{0.7589}   & 27.00 & 0.7178  & \textcolor{blue}{24.68} & \textcolor{blue}{0.7347}                  \\
					& Proposed   & \textcolor{red}{30.94} & \textcolor{red}{0.8701}   & \textcolor{red}{27.82} & \textcolor{red}{0.7597}   & \textcolor{red}{27.04}  & \textcolor{red}{0.7186}  & \textcolor{red}{24.73}   & \textcolor{red}{0.7364}         \\
					\midrule[0.8pt]
					\multirow{9}{*}{RCAN-B}& Vanilla     & 31.24 & 0.8752  & 27.99 & 0.7640  & 24.17 & 0.7231  & 24.94 & 0.7459          \\
					& FitNet~\cite{Fitnet}   & 31.35  & 0.8775   & 28.07 & 0.7669  & 24.22  & 0.7248   & 25.06 & 0.7507     \\
					& AT~\cite{AT}   & 31.26  & 0.8758  & 28.02 & 0.7648  & 27.20 & 0.7237  & 25.01  & 0.7490         \\
					& VID~\cite{VID}   & 31.32  & 0.8764 & 28.07 & 0.7657  & 27.21 & 0.7241  & 25.04  & 0.7501                  \\
					& FAKD~\cite{SA}   & 31.31  & 0.8767  & 28.03  & 0.7651   & 27.19  & 0.7241  & 25.00  & 0.7485                 \\
					& ICKD~\cite{ICKD} & 31.27 & 0.8760  & 28.02 & 0.7650   & 27.19 & 0.7237  & 25.01 & 0.7489    \\
					& L2T-ww\tnote{*}~\cite{L2T-ww}   & 31.39 & 0.8784	& 28.08 & 0.7664   & 27.22  & 0.7248   & 25.09  & 0.7516               \\
					& PISR\tnote{**}~\cite{PISR}   & \textcolor{blue}{31.40} & \textcolor{blue}{0.8785}	& \textcolor{blue}{20.10}  & \textcolor{blue}{0.7664}   & \textcolor{blue}{27.23} & \textcolor{blue}{0.7250}  & \textcolor{blue}{25.11} & \textcolor{blue}{0.7528}              \\
					& Proposed   & \textcolor{red}{31.47} &	\textcolor{red}{0.8798} & \textcolor{red}{28.16}  & \textcolor{red}{0.7680}  & \textcolor{red}{27.27}  & \textcolor{red}{0.7264} & \textcolor{red}{25.19}   & \textcolor{red}{0.7562}        \\
					\midrule[0.8pt]
					\multirow{9}{*}{RCAN-C}& Vanilla     & 31.77 & 0.8841  & 28.31 & 0.7724  & 27.39 & 0.7305  & 25.51  & 0.7674       \\
					& FitNet~\cite{Fitnet}   & 31.94  & 0.8870 & 28.38  & 0.7740 & 27.43  & 0.7319  & 25.61  & 0.7710     \\
					& AT~\cite{AT}   & 31.85  & 0.8860 & 28.36 & 0.7737  & 27.41 & 0.7312  & 25.58  & 0.7696       \\
					& VID~\cite{VID}   & 31.89  & 0.8861  & 28.38 & 0.7740  & 27.42 & 0.7314  & 25.59  & 0.7701        \\
					& FAKD~\cite{SA}   & 31.89  & 0.8865  & 28.38  & 0.7741  & 27.43  & 0.7319   & 25.57  & 0.7696             \\
					& ICKD~\cite{ICKD} & 31.85  & 0.8854  & 28.34  & 0.7732  & 27.40 & 0.7309  & 25.55  & 0.7690    \\
					& L2T-ww\tnote{*}~\cite{L2T-ww}   & 31.91   & 0.8862 	& 28.37  & 0.7739   & 27.42   & 0.7316   & 25.61   & 0.7704      \\
					& PISR\tnote{**}~\cite{PISR}   & \textcolor{blue}{31.96} & \textcolor{red}{0.8873}	& \textcolor{blue}{28.40} & \textcolor{red}{0.7749}   & \textcolor{blue}{27.45} & \textcolor{blue}{0.7321}  & \textcolor{blue}{25.65} & \textcolor{blue}{0.7720}              \\
					& Proposed   & \textcolor{red}{31.96} & \textcolor{blue}{0.8871}  & \textcolor{red}{28.40} & \textcolor{blue}{0.7746}  &\textcolor{red}{27.45} & \textcolor{red}{0.7325}   & \textcolor{red}{25.71}  & \textcolor{red}{0.7742}       \\
					\bottomrule[1pt]
				\end{tabular}
				\begin{tablenotes}
					\footnotesize
					\item[*] For a fair comparison, L2T-ww is settled manually the same distillation position as other methods do, rather than via meta-learning.
					\item[**] Except that PISR introduces privileged information to train a specific teacher network, all other distillation schemes use the RCAN-T as the teacher.
				\end{tablenotes}
			\end{threeparttable}
		\end{small}
	}
\end{table*}

\begin{figure*}[htbp]
	\begin{center}
		\includegraphics[width=0.92\linewidth]{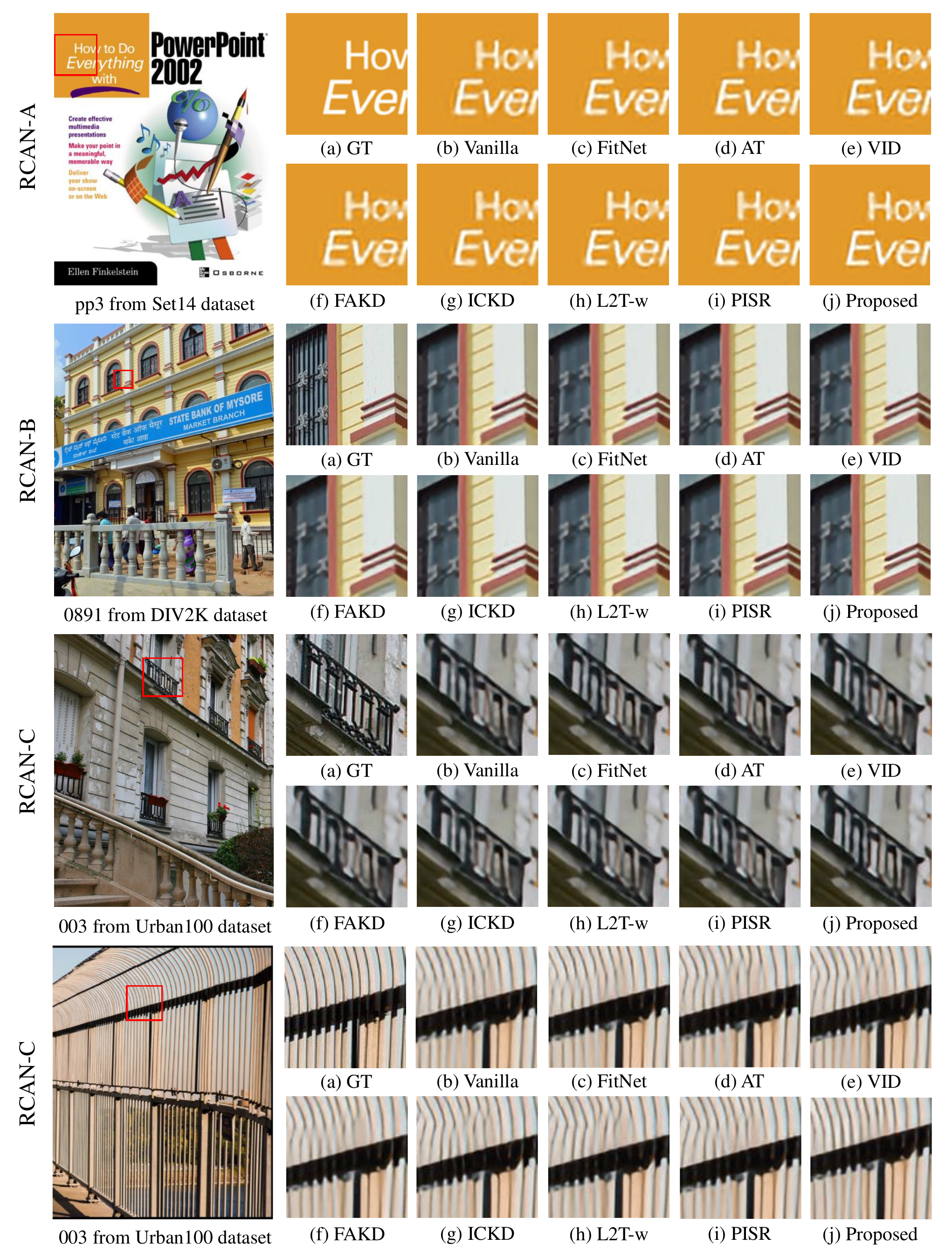}
	\end{center}
	\caption{Visual comparison of different knowledge representation on img046 from Urban100~\cite{Urban100} dataset: (a) w/o distillation, (b) FitNet~\cite{Fitnet}, (c) AT~\cite{AT}, (d) SA~\cite{SA}, (e)L2T-ww~\cite{L2T-ww} and (f) the proposed MKD.}
	\label{fig:long}
	\label{fig:Comparison_visual}
\end{figure*}

\subsection{Comparison results}

We compare the proposed meta knowledge distillation method with some of the existing knowledge representation and super-resolution related distillation algorithms, including FitNet~\cite{Fitnet}, AT~\cite{AT}, VID~\cite{VID}, FAKD~\cite{SA}, ICKD~\cite{ICKD}, L2T-ww~\cite{L2T-ww} and PISR~\cite{PISR}. Among them, AT, FAKD and ICKD are artificially defined representations of knowledge. The others directly employ feature maps of the teacher's intermediate layers as the transferred knowledge and then use learnable parameters to make students imitate the teacher. FAKD and PISR are proposed for super-resolution task. Since the manually defined methods usually employ normalization operations to process feature maps, if the corresponding distillation loss is used as a regularization term for the original reconstruction loss, their impact on $\mathcal{L}_{org}$ (L1 loss) will be very marginal. To ensure the fairness of performence comparison, the stage-wise training procedure proposed by FitNet is applied to all of these KD methods. In addition, the focus of this paper is mainly on the representation of knowledge, so four distillation positions are fixed, and the weighted network determining where to transfer in L2T-ww is removed in the experiment. Except that PISR introduces privileged information to train a dedicated teacher and initializes the student with the parameters of the teacher, the configuration of students and teacher is still in accordance with Table~\ref{tab:config}, and the training details follow the description in section~\ref{subsec:labeltwo}.

\begin{table*}[!h]
	\centering
	\caption{The results of implement the proposed meta knowledge distillation on typical SR structures.}
	\label{tab:implement}
	\setlength{\tabcolsep}{3mm}{
		\begin{small}
			\begin{threeparttable}
				\begin{tabular}{c|c|cc|cccccccc}
					\toprule[1pt]
					Network	& Type  & Param. & Flops  & Set5~\cite{Set5} & Set14~\cite{Set14} & BSD100~\cite{B100} & Urban100~\cite{Urban100}                 \\
					\midrule[0.8pt]
					\multirow{3}{*}{EDSR~\cite{EDSR}}& Teacher     &43.1M &13179.19G & 32.48 / 0.8955  & 28.82 / 0.7860  & 27.72 / 0.7425  & 26.65 / 0.8036      \\
					& Student(Vanilla)  &40.3K  &19.58G  & 30.61 / 0.8632  & 27.59 / 0.7542  & 26.92 / 0.7143  & 24.49 / 0.7264    \\
					& Student(Proposed)  &40.3K  &19.58G  & 30.84 / 0.8681  & 27.74 / 0.7575  & 27.01 / 0.7172  & 24.65 / 0.7332      \\
					\midrule[0.8pt]
					\multirow{3}{*}{RDN~\cite{RDN}}& Teacher\tnote{*}  & 22.3M  & 5961.20G  & 32.29 / 0.8941  & 28.71 / 0.7836  & 27.68 / 0.7410  & 26.51 / 0.7998          \\
					& Student(Vanilla)  & 122.7K  & 41.18G  & 31.12 / 0.8741  & 27.94 / 0.7632  & 27.13 / 0.7220  & 24.90 / 0.7440          \\
					& Student(Proposed) & 122.7K  & 41.18G  & 31.38 / 0.8787  & 28.11 / 0.7674  & 27.24 / 0.7252  & 25.12 / 0.7527   \\
					\midrule[0.8pt]
					\multirow{3}{*}{SAN~\cite{SAN}}& Teacher  & 15.8M  & 4268.24G & 32.64 / 0.8971   & 28.91 / 0.7871  & 27.77 / 0.7438     & 26.83 / 0.8079        \\
					& Student(Vanilla)  & 48.0K   & 21.48G  & 30.65 / 0.8648   & 27.61 / 0.7549  & 26.95 / 0.7156   & 24.54 / 0.7285            \\
					& Student(Proposed)   & 48.0K  & 21.48G   & 30.94 / 0.8697   & 27.83 / 0.7592   & 27.07 / 0.7193   & 24.77 / 0.7379     \\
					\midrule[0.8pt]
					\multirow{3}{*}{HAN~\cite{HAN}}& Teacher  & 16.1M   & 4309.49G	& 32.60 / 0.8963   & 28.88 / 0.7870  & 27.77 / 0.7439   & 26.78 / 0.8070         \\
					& Student(Vanilla)  & 123.9K   & 41.03G	& 31.32 / 0.8768   & 28.02 / 0.7650  & 27.20 / 0.7244    & 25.04 / 0.7499         \\
					& Student(Proposed)   & 123.9K   & 41.03G   & 31.51 / 0.8802   & 28.14 / 0.7676   & 23.27 / 0.7264    & 25.18 / 0.7555      \\
					\bottomrule[1pt]
				\end{tabular}
				\begin{tablenotes}
					\footnotesize
					\item[*] We trained the teacher of RDN from scratch, while the others are obtained from their official Pytorch pre-training weights.
				\end{tablenotes}
			\end{threeparttable}
		\end{small}
	}
\end{table*}

Comparison results are shown in Table~\ref{tab:analysis} and the relationship between complexity and performance is drawn in Figure~\ref{fig:Tradeoff} for more straightforward observation. In general, when the gap between students and teachers is enormous, especially when the number of channels is significantly different, knowledge representations with learnable parameters outperform artificially defined ones. They are more flexible for applications with different gaps in knowledge distillation. Among the artificially defined representations, for RCAN-A with the most enormous gap among the student settings, the inter-channel correlation in ICKD~\cite{ICKD} outperforms the spatial attention in AT and the spatial affinity in FAKD~\cite{SA}. Conversely, for RCAN-C, the loss of channel information in knowledge representation is less affected than the loss of spatial information. This fact is also consistent with the cognition of super-resolution task and shows differences in the guidance acquired by students with different complexities. In addition, these requirements are likely to change with the optimization process. Hence, it is difficult to find a one-size-fits-all manual definition method that can handle all the needs based on experience and prior knowledge. Compared with FitNet~\cite{Fitnet}, L2T-ww~\cite{L2T-ww} determines the weight of each feature map in $\mathcal{L}_{kd}$ through meta-learning so that the student network can pay more attention to the similarity of more essential features. This strategy obviously affects RCAN-A, which shows that the features extracted by convolution layers are not equally important. Still, the knowledge screening of L2T-ww only stays at the feature level without further conversion. Hence, when the gap between students and teachers decreases, the effect of weighted feature distillation is no longer significant. Own to introducing the privileged information to train specific teacher models, PISR~\cite{PISR} performs significantly better than VID~\cite{VID}. However, in terms of knowledge representation, the same strategy as VID is still adopted, that is, to maximize the mutual information between students and teachers through variational inference without discriminating the features extracted from teachers.

In contrast, we propose a more flexible approach to determine the representation of knowledge so as to teach students according to their aptitude. This method not only achieves the best performance on the most metrics but can also steadily improve students' performance with different complexity. Due to setting L1 loss as the meta objective, the behavior on PSNR is slightly more significant than that of SSIM. In addition, with the texture-aware kernels, the method performs better on regularly textured and dense images, such as the Urban100 dataset.
The advantage is also reflected in the subjective comparison. As shown in Figure~\ref{fig:Comparison_visual}, compared with other distillation schemes, the students optimized by the proposed method can restore clearer and sharper edges, such as text and building wall lines, and can effectively suppress artifacts such as ringing effects. More significantly, it can effectively alleviate the problems of dislocation and geometric distortion that are likely to occur in dense texture reconstruction, making the reconstruction results more realistic and reliable.

\subsection{Implement on more typical SR structures}

Since the proposed knowledge distillation method has no concern with the super-resolution network architecture, we apply it to more super-resolution algorithms, including EDSR~\cite{EDSR}, RDN~\cite{RDN}, SAN~\cite{SAN} and HAN~\cite{HAN}. Except that we trained the teacher of RDN from scratch, while the others are obtained from their official Pytorch pre-trained weights. As in the previous settings, the student networks are acquired by reducing the network depth and the number of convolution channels, and the Flops are calculated on an LR image of size $512\times512$. The experimental results are shown in Table~\ref{tab:implement}. The proposed meta knowledge distillation method on different network structures can effectively improve the reconstruction quality compared to training from scratch. Meanwhile, it also shows that the proposed algorithm can be adapted to various operators in the CNN-based super-resolution network, including dense connection, attention mechanisms, and non-local operations. Experiments conducted on different SISR datasets demonstrate that our proposed method can achieve advanced performance in the single image super-resolution task.

\section{Conclusion}
\label{sec:conclusion}

Knowledge distillation (KD) has demonstrated its advantages in some vision applications, and it is also worth investigating its application in super-resolution task. 
In this paper, we propose a model-agnostic meta knowledge distillation method for the single image super-resolution task, from the perspective of knowledge representation. Different from the existing KD algorithms, the proposed method takes full consideration of students' abilities and provides a more flexible way to determine the transferred and digested knowledge. KRNets are designed to convert the intermediate features of teachers and students into knowledge through learnable parameters, and then optimize these parameters through meta-learning so that the students are optimized to improve their reconstruction quality rather than just imitating the teachers. Texture-aware kernels are generated to focus on more helpful information and facilitate the reconstruction of detailed textures. Experiments conducted on SISR benchmarks demonstrate that the proposed method is superior to other knowledge distillation algorithms and can significantly improve the reconstruction quality of super-resolution networks without adding any computational complexity.

\bibliography{ms}

\end{document}